# Fuzzy Ontology-based Sentiment Analysis of Transportation and City Feature Reviews for Safe Traveling

Farman Ali[1], Daehan Kwak[2], Pervez Khan[3], S. M. Riazul Islam[1], Kye Hyun Kim[1], K. S. Kwak[1*]

[1]Inha University, South Korea
[2]Rutgers University, USA
[3]Incheon National University, South Korea.
E-mail: farmankanju@gmail.com; pervaizkanju@hotmail.com; riaz@inha.ac.kr; kwakno1@cs.rutgers.edu, kyehyun@inha.ac.kr
*Corresponding author: K. S. Kwak (e-mail: kskwak@inha.ac.kr)

**Abstract:** Traffic congestion is rapidly increasing in urban areas, particularly in mega cities. To date, there exist a few sensor network based systems to address this problem. However, these techniques are not suitable enough in terms of monitoring an entire transportation system and delivering emergency services when needed. These techniques require real-time data and intelligent ways to quickly determine traffic activity from useful information. In addition, these existing systems and websites on city transportation and travel rely on rating scores for different factors (e.g., safety, low crime rate, cleanliness, etc.). These rating scores are not efficient enough to deliver precise information, whereas reviews or tweets are significant, because they help travelers and transportation administrators to know about each aspect of the city. However, it is difficult for travelers to read, and for transportation systems to process, all reviews and tweets to obtain expressive sentiments regarding the needs of the city. The optimum solution for this kind of problem is analyzing the information available on social network platforms and performing sentiment analysis. On the other hand, crisp ontology-based frameworks cannot extract blurred information from tweets and reviews; therefore, they produce inadequate results. In this regard, this paper proposes fuzzy ontology-based sentiment analysis and semantic web rule language (SWRL) rule-based decision-making to monitor transportation activities (accidents, vehicles, street conditions, traffic volume, etc.) and to make a city- feature polarity map for travelers. This system retrieves reviews and tweets related to city features and transportation activities. The feature opinions are extracted from these retrieved data, and then fuzzy ontology is used to determine the transportation and city-feature polarity. A fuzzy ontology and an intelligent system prototype are developed by using Protégé web ontology language (OWL) and Java, respectively. The experimental results show satisfactory improvement in tweet and review analysis and opinion mining.

## 1. Introduction

Traffic monitoring and transportation facilities are two major concerns of intelligent transportation systems (ITSs). At present, sensor network based transportation systems are employed to monitor the entire traffic network and to provide emergency services in response to associated activities occurring on the roads. In addition, people use various devices in vehicles, such as small-scale collision radars, global positioning systems, radio transceivers, and sensing devices to retrieve real-time information in order to travel safely. Most of these systems are designed for short-range communications and are unable to collect all traffic-related data. They also fail to notify drivers before entering risk zones, resulting in wasted time and consumption of extra fuel. As a solution, precise information is needed that can be utilized to rapidly identify risky spots and activities. However, because of the wonderful trends in interpersonal organization during the most recent decade, network platforms, like the various community blogs, Facebook, Twitter, and online forums have turned into the richest sources of real-time data (Liu et al., 2005; Cao et al., 2014; Kim et al., 2006).

Currently, people utilize social media sites to share their opinions on different issues associated with transportation (e.g., traffic collisions, traffic jams, and landslides or rockslides). New clients come across the reviews of others and react to them regarding the same subject (e.g., roads or city streets jammed, street-side organizations, and associations). Conversely, a large volume of tweets or reviews can puzzle web surfers trying to determine immediate and safe routes. In most cases, people give their assessments about transportation in terms of features like "the information of the traffic police officer was extremely definite and helpful"; or "at least three hurt in a two-vehicle mishap close to the downtown"; or "Victoria downtown has a ton of offices, yet the street is jammed". Opinion reviews are concealed in forums and blogs, making it difficult for the user to extract meaningful information. Sentiment-analysis or opinion mining is the process of extracting valuable information from public tweets and reviews about a specific topic using text analysis methods and natural language processing (Ali et al., 2015). Therefore, sentiment analysis can help ITSs to improve traffic monitoring and transportation facilities.

Most of the present systems are not good enough to categorize the correct positive and negative sentiment words and to recognize terms for the degree of feature polarity. It is critical to determine whether a tweet or review represents a strongly positive, positive, neutral, negative, or strongly negative polarity. In addition, data extraction frameworks primarily take classic ontologies into account. The classic ontology addresses only crisp data, and cannot recover alluring results from the cloudy wellspring of social network data. Different systems and sites on transportation and travel give a rating score to a city, taking rates of various variables into account, for example, green space, cleanliness, tourism offices (restaurants, hotels, parks, and so on.), safety, peacefulness, low crime rates, and so on (Gilboa et al., 2015). These rating scores do not convey precise information, in spite of the fact that tweets or reviews are significant because they offer some assistance to the transportation administrators and travelers so they will know about all positive and negative points of each feature. However, it is troublesome for ITSs and travelers to peruse every tweet and review and come to an important conclusion in regards to their necessities from city factors. For the most part, people shroud their feelings about a city, rather examining it in terms of city-features; for instance, "the marvelous foggy weather of our excellent Manhattan" (Bertrand et al., 2013). Therefore, it is critical to locate the individual city-feature polarity, and thereafter, the overall city polarity. On that, this paper presents fuzzy ontology-based sentiment analysis and semantic web rule language (SWRL) rule based decision-making knowledge that will help ITSs and travelers. Below are the main contributions of this research.

- The proposed system effectively retrieves reviews and tweets related to city features (e.g., bus and train stations, bridges, parks, restaurants, airports, medical centers, and hotels) and transportation activities (e.g., collisions, roadsides, congestions, and traffic jams).
- An unsupervised linear method is used to automatically extract related sentiments from online consumer tweets and reviews.
- Fuzzy ontology-based semantic knowledge is developed to extract individual transportation activities and city features with the opinion words. Then, fuzzy ontology is used to compute the polarity of transportation activities, city features and the overall city from the degrees of the terms (strongly negative, negative, neutral, positive, and strongly positive).
- SWRL rule based decision-making knowledge is proposed to find the major causes of traffic congestion and negative polarity.
- An opinion map of the city features and city transportation activities is automatically designed by using semantic knowledge and sentiment analysis results, which enrich the performance of ITSs by determining the traffic problems and providing safe routes for travelers.

The remainder of this paper is organized as follows. Section 2 provides the state-of-the-art work in this area. Section 3 illustrates the basic concept of the proposed system architecture. Fuzzy ontology based semantic knowledge is explained in Section 4. Section 5 briefly explains the overall scenario and internal process of the sentiment analysis and polarity computation. Section 6 presents the experimental results.

## 2. Related works

Transportation data processing and useful traffic information extraction from social networks are hot topics in intelligent transportation systems. Escalation in the amount of traffic-related data on the internet has made the task of ITSs more challenging. Presently, ITSs monitor every corner of the transportation system within range of a sensor network. These sensor network based systems cannot collect precise information related to traffic. In addition, people use small-scale devices in vehicles to retrieve information to travel safely. These devices can collect short-range information but are unable to retrieve all traffic-related data. As a solution, intelligent work is needed to automatically extract and compute traffic data and provide meaningful information. This can improve transportation safety, can solve traffic congestion problems in big cities, and can provide comfortable traveling in urban areas. A variety of research has been done on this issue, and several ideas have been proposed for vehicular network problems (Whaiduzzaman et al., 2014).

Yu et al. (2013) proposed a hierarchical cloud architecture for vehicular networks. The main aim of this system is to design a ubiquitous cloud environment for mobile vehicles using various resources within ITS infrastructures, such as vehicles, data centers and roadside units. The system uses a three-layer architecture that allows vehicles to access the cloud services, provides satisfactory communications quality, and maintains the efficiency of the cloud services. A parallel hybrid genetic algorithm was proposed to update a vehicle's place and time information (Nijanthan et al., 2015). In this research, the authors compared their system with conventional cloud computing and mentioned different research issues and future directions. The authors addressed security and privacy issues in the vehicular cloud concept to allow only authenticated users to access the system. The security protocol validates a sender by using a digital signature under the existing architecture. Naturalistic observation was used to define an overtaking vehicle's trajectory (Papakostopoulos et al., 2015). The system compares clearance margins between an overtaking vehicle and the overtaken vehicle at the end of the maneuver. It examines this situation in two ways. First, the system determines whether a collision risk occurs at the end of the maneuver between the oncoming and the overtaken vehicles, and it then finds the balance between the vehicles to check whether the appearance of the vehicle is affected or not while overtaking.

A cloud–based service framework was developed to conserve energy in an intelligent transportation system (Hsu et al., 2015). In this framework, a massive storage facility was designed for transportation information, which avoids the use of a hard disk at the user site and improves reliability. Mobile vision techniques are used to recognize real-time traffic conditions. This technique is based on road-side camera data. These data are used to develop fuel-efficient route navigation technology to save time and reduce fuel consumption. Various approaches have been proposed about transportation modeling challenges (Vlahogianni, 2015). One approach presented computational intelligence algorithms and deductive data analyses to extract useful patterns from big data in transportation, which increases flexibility and accuracy, and handles uncertain changes in big data. A leap frog method was proposed to improve the performance of a serial map-matching algorithm for a transportation data center, which preserves location privacy and maintains accuracy (Huang et al., 2015). This system uses map-matching in the public cloud, while the travel paths are used within an organization's private cloud to automatically arrange data segregation and location according to data privacy specifications. The NeverStop idea was proposed to utilize genetic algorithms and fuzzy logic in big data ITSs (Wang et al., 2016). It uses radio frequency identification (RFID) devices to retrieve information on vehicles that pass a green traffic light, and those that wait for at a red light, and stores this information in EBOX II devices in a central server for further monitoring. Botta et al. (2016) presented new challenges and research issues in the cloud's Internet of Things (IoT) paradigm. In this literature survey, complementary details are discussed on the integration of cloud services and the IoT. Various available platforms are debated, and challenges are examined in detail to present recent problems in the research.

During the past few years, information extraction and sentiment analysis have been used to facilitate the public sectors. Traffic sentiment analysis based on social data was proposed to meet the needs of safety and information exchange in ITSs (Cao et al., 2014). This existing work presented a step-by-step procedure to develop related bases for traffic sentiment analysis, proposing sentiment polarity computation algorithms using a rules base, comparing the good and bad aspects of rule-based and learning-based approaches. Sentiment analysis in the Twitter social network was used to predict an influx of tourists to Peru (Linares et al., 2015). The proposed system uses a web crawler to retrieve tweets, and a naïve Bayes algorithm to classify them. A real-time prototype was implemented to perform sentiment analysis for tweet and retweeting structures (Lin et al. 2014). This system uses a rule-mining algorithm to efficiently mine the sentiments for real-world Weibo posting. In addition, a microblog-oriented lexicon is employed to deal with the precision of emotions and online languages. Sentiment analysis based on machine learning methods was proposed to identify tourist attraction targets (Lin et al., 2010). Both document and sentence-level opinions are considered, and the target opinions are determined after comparing all of them. However, document and sentence-level based sentiment cannot efficiently classify the users' comments to find their real feelings. Pipelines and joint models are trained by using a large-margin prediction method to find the sentiment expressions in texts and to declare the polarity of the extracted expressions (Johansson et al., 2011). This model explains the extraction of integrating opinions and the classification of polarity using the features' global polarity structures. The authors compare the proposed system with a previously published one using the Multi-Perspective Question Answering (MPQA) Opinion Corpus, and achieved significant improvement in 13 absolute points in a function measure. Online social media data were used to measure rider sentiments for transit rider satisfaction evaluation (Collins et al., 2013). The proposed system used the metrics to evaluate timeliness, safety, cleanliness, and ridership. Sentiment analysis accurately classified rider opinion over a period of time and showed the results in the form of positive, negative, and normal sentiments. Opinion mining of riders' tweets was employed to find the popularity of bike-sharing in big cities (Das et al., 2015), also known as smart bike and capital bike share. This system deeply explained the extraction procedure of tweets and information from websites. Intelligent topical sentiment analysis was used to classify E-Learners (Ravichandran et al., 2015). The authors applied unsupervised algorithms Sentence level opinion classification, which is different from traditional document-level classification. It significantly improved the performance using bigrams, compared with a supervised algorithm. A sentiment analysis based system was developed to easily understand the trends in citizen behavior and their sentiments regarding city services and events (Villena-Roman et al., 2014). After identifying the negative sentiments, this system automatically alert the emergency system to improve efficiency.

Recently, fuzzy logic and ontologies have been used in the field of transportation knowledge management, sentiment analysis, information retrieval, and information extraction. An ontology is the proper modification of a shared conceptualization in a specific domain, which is human-understandable and machine-readable format (Ali et al., 2015). An ontology was used to develop a traffic accidents management system (Yue et al., 2009). This ontology captured valuable information about pedestrians, climate, environments, and roads, which is necessary for traffic-safety situations. However, various organizations have developed traffic management systems according their needs, but there is a limited amount of fuzzy logic-based semantic knowledge for information retrieval and expression. Ali et al. (2016) proposed a fuzzy ontology-based opinion mining system to monitor the transportation network. This existing system computes the polarity of 6 features, which is not enough to monitor the entire transportation activities and city features. It does not provide the detail process of assigning initial values to the corresponding opinion words and the detail process of integrating the extracted opinionated phrases and their polarity values to verify the feature polarity. In addition, it uses simple web crawler and single keywords-based query to retrieve information. However, it is difficult to collect meaningful data about a specific topic, because most of the data are expressed in different ways, and the context of the writing is disorganized. Therefore, we proposed a fuzzy ontology-based crawler to collect data from vague sources. Further, the existing system employs resource description framework (RDF) datasets to classify the polarity. Also, sentiment analysis needs both SentiWordNet-based raw datasets along with RDF datasets to increase the precision and accuracy of polarity. Therefore, we proposed a fuzzy ontology, fuzzy inference layer, and support vector machine-based system to overcome the limitation of the existing system. RahmathP et al. (2014) proposed fine-grained opinion classification of online product reviews. This system uses unsupervised machine learning and fuzzy linguistic modifiers for sentiment explanation. The fuzzy logic of sentiment words gives better accuracy than other approaches. However, this approach needed ontology-based smart semantic knowledge to determine the linguistic variable for product feature polarity. Sentiment analysis of product review documents was classified by oriented fuzzy logic (Indhuja et al., 2014; kawatheker and Kshirsagar, 2012). These systems classify movie reviews, emotions, and instance. Fuzzy logic deals with any proposition and computes the polarity value of documents, and combines them in an opinion classification task. A fuzzy and semantic approach was proposed to mine online reviews (Al-Maimani et al., 2014). This system solved the semantic problem in sentiment analysis. The fuzzy logic improved the extraction and summarization of sentiments with their strengths and weight. A mathematical model of information generation on websites was developed for sentiment analysis (Trung et al., 2014). This decision support system finds the interaction between opinion words and online information, and then uses these opinion words in an efficient way to increase the effect of information generation. A classic ontology was employed to solve the feature classification problem in the domain of online reviews (Zhao and Li, 2009; Lau et al., 2009). This classic ontology is useful for extracting data from organized information. However, most tweets and reviews on social networks are unstructured or in blurred format. Therefore, the mentioned classic ontology cannot define feature fuzzy terms (e.g., traffic {jammed, fast, slow, medium and heavy}). The integration of fuzzy logic with a classic ontology works excellently during uncertain input execution.

The discussion of previous studies revealed that most of the existing research is flawed in the areas of transportation data processing and traffic network monitoring. Most proposed traffic-monitoring systems are based on a sensor network. Unfortunately, sensor network based systems are unable retrieve all traffic-related data. In addition, a classic ontology is incapable of extracting data from imprecise information, and to perfectly address intensively blurred data related to transportation activities. Therefore, this proposed fuzzy ontology based semantic knowledge is a novel effort to automatically extract and process traffic data and provide useful information.

## 3. The proposed system architecture

This proposed architecture is based on a fuzzy logic and SWRL rule-based ontology to deal with any type of real scenario related to city and transportation reviews and tweets. Fig. 1 shows the architecture of the proposed system. In this proposed framework, the social network data will go through many phases in order to compute polarity as follows.

- In the data collection phase, this system uses the application programming interfaces (APIs) of e-commerce sites to retrieve reviews and tweets from social network sites (such as Twitter, Facebook, news, announcements, and traveling sites) for appropriate city and transportation activities. Tweets can be retrieved using representational state transfer (REST) APIs and Streaming APIs. Using REST APIs, users can retrieve 3,200 of the most recent tweets. The REST APIs allow users to employ queries for tweets retrieval. These queries contain a set of keywords, operators (AND and OR), radius and centroid (e.g., accident AND (car OR vehicle)). The limits of REST APIs include 350 queries per 15 minutes for single user and 3200 tweets per query. Different queries are designed to retrieve traffic related tweets. These queries are based on 500 keywords related transportation and city. Some usual queries related to transportation are as follows. Query 1: "traffic AND accident, car AND collision, vehicle OR Car OR bus AND crash, car AND accident, vehicle AND collision, car AND crash, bus AND accident". Query 2: "road closed AND traffic congestion OR accident, slow traffic AND accident, traffic jammed AND accident". Query

3: "road AND traffic collision OR accident, street AND vehicle accident, traffic jam AND accident". Query 4: "traffic light AND accident, traffic light AND street AND accident, traffic AND highway light AND accident". While reviews from Facebook are manually collected because for every feature there is a separate Facebook places page. The data collection from news forums related to transportation activities is easy and valuable, but it is difficult to extract opinions from these data. Twitter, Facebook, and other traveling sites contain discussions on specific topics, which can easily be used for polarity computation. However, it is difficult to collect meaningful data about a specific topic, because most of the data are expressed in different ways, and the context of the writing is disorganized. Therefore, a fuzzy ontology-based crawler is used to collect data from vague sources. This crawler is restricted to pursuing appropriate links that hold transportation activities and city feature reviews and tweets.

- In the pre-processing phase, text collection and corpus pre-processing are the main objectives before sentiment mining. This phase converts the collected data into sentiment clues to easily extract features and sentiment words. Tweets and reviews are pre-processed to remove needless content, such as special symbols (#, @, dates, etc.), and articles (a, an, the), and the text is stemmed to reduce words to their roots, such as cleaned, cleaner, cleaned to the root word clean. Further, the text is tagged to identify the words' parts of speech (POS). The tweets and reviews are split into sentences, which are then confirmed as complete clauses with nouns and a verb phrase. The splitting of this text makes fuzzy ontology-based feature tagging more easy.
- In the data cleaning phase, various queries are used to collect particular tweets and reviews from hazy social networks. These queries include different phrases with a city name, such as car accident on the highway or road or street; vehicle collision or crash; traffic jam on road or highway with an accident; highway or road or freeway and accident; slow traffic on highway or road and road closed and accident. The opinion mining or sentiment analysis dataset is prepared to retrieve those tweets and reviews that can describe the dataset topics.
- After pre-processing and data cleaning, the proposed system extracts the transportation activities and city features with sentiment words from tweets and reviews. This extraction pairs feature words with sentiment words. Subjects and objects are mostly extracted by text mining and document mining (Cao et al., 2014). A subjective set contains positive and negative sets of opinion words. Previous studies suggest that common and proper nouns are mostly employed as object, whereas personal pronouns are used as subject. People use the simple past tense to express their ideas as subjective texts for themselves and objective texts for others. Both adjectives and adverbs are used to express sentiments or opinions. However, superlative adjectives are rarely used while comparative adjectives are mostly used in tweets and reviews. Another dataset for positive text detection is superlative adverbs (quickest, fastest, longest, etc.) which shows a positive opinion. The past tense verbs in negative sets express negative opinions, because users always employ it to express a negative opinion regarding their dissatisfaction (died, jammed, lost, etc.). These three datasets (positive, neutral, and negative) with an n-gram technique are utilized to extract features with an opinionated phrase. The n-gram technique as a binary feature may extract a feature efficiently, but it may not be useful to indicate the overall opinion. Therefore, the proposed fuzzy ontology based semantic knowledge identifies features in a sentence and computes their polarity. Every extracted feature from tweets and reviews is checked with the mentioned features in the ontology, and only validated feature sentences are computed. Then, the opinion aggregation formula is used to compute the sentiment word scores of features, $f_i$, in sentence $s_k$ (Ding et al., 2007).
- The fifth phase is SentiWordNet scoring. The SentiWordNet database contains three kinds of score for each opinionated word: positive, neutral, and negative. It is employed to compute each sentiment word score (Sayeedunnissa et al., 2013). If the score of a sentiment word is not present in the SentiWordNet database, then a zero is assigned to the opinion phrase. For example, in the term 'big-city', big is a sentiment word, and its score is 0.75. In order to compute the effect of linguistic fuzziness on sentiment word, fuzzy logic with the SentiWordNet tool and ontology is used. This method combines the sentiment words of each feature in a tweet or review sentence and determines the polarity value using a fuzzy logic technique. The polarity computation process is explained in Section 5.
- The sixth phase is an opinion lexicon. Opinion mining or sentiment analysis is mostly dependent on two associated components: an opinion lexicon, and polarity computation of opinion words. The opinion lexicon is necessary for accurate sentiment analysis. Therefore, a sentiment lexicon based on positive and negative opinion words is employed for sentiment analysis (Liu, 2010).

Fuzzy ontology, SWRL rules, polarity computation and sentiment analysis of transportation and city feature are more deeply explained in the next sections.

## 4. Fuzzy ontology-based semantic knowledge

An ontology is shared knowledge of a specific domain among people and systems. It is written in a specific language called a web ontology language (OWL). To achieve efficiency for the proposed ontology, a classic ontology was designed using Protégé OWL. A fuzzy OWL plug-in was then employed to convert the classic ontology into a fuzzy ontology (Bobillo et al., 2011). The fuzzy OWL plug-in is used to assert fuzzy terms in the ontology. The classes, instances, concepts, and axioms of a classic and a fuzzy ontology are the same. However, all the concept values of a classic ontology are blurry terms. A classic ontology cannot handle uncertainty. A fuzzy ontology is generally defined to express vague knowledge using fuzzy concepts. Therefore, this system needs fuzzy ontology based semantic knowledge to handle any kind of situation related to sentiment analysis. Useful transportation knowledge is accumulated regarding traffic, accidents, roads, vehicles, climate, and the environment, and is delivered to the ontology for a traffic accident management system, as shown in Fig 2. Mathematically, an ontology can be defined as follows:

$$\widetilde{Ontology} = (C, P, R, V, V_c) \qquad (1)$$

In the above equation, the notations C, P, R, V, and $V_c$ stand for concepts, properties, relationships of classes, values, and constraint values of properties, respectively (Ali et al., 2015). Zadeh, (1965) presented the concept of a fuzzy set in 1965. Fuzzy set theory describes vague boundaries, such as negative, neutral and positive. A fuzzy set, F, over the universe of discourse A can be shown by its membership function $\mu_F$, which presents an element 'A' in the interval [0, 1] (Ali et al., 2015).

$$\mu_F(A): A' \to [0, 1] \qquad (2)$$

In the above equation, A' belongs to A and $\mu_F$ presents the membership degree by which A' ∈ A. A' is considered a full member of set A if $\mu_F(A) = 1$. A is considered a partial member of set A if $\mu_F(A)$ is between zero and one (e.g., 0.63) (Ali et al., 2015).

A fuzzy ontology exchanges the knowledge among feature extractions (Jeong et al., 2011), reviews classifications, and performs feature polarity identification. Therefore, the description of polarity for feature classification using a fuzzy ontology expedites the proposed transportation sentiment- analysis system. This system categorizes the features and extracts the correct feature polarity terms. The fuzzy ontology defines the concepts of transportation activities and city-feature polarity generation. The OWLViz plugin of Protégé is used to automatically generate the graphical representation of all features as shown in Fig. 2. In fig. 2, the arrow direction shows that every subclass is a feature of upper class such as 'closed' and 'slow' are subclasses or features of road and traffic, respectively. These features are used to find the major causes of upper class negative polarity. For example, if the polarity of traffic is strong negative, then subclasses (Jammed, slow, traffic collision, and heavy) can be employed to find the cause of traffic polarity. These classes along with their polarity terms are used to make a SWRL rules for polarity computation as shown in table 1.

(Fig. 2 is placed on page 17)

(Table 1 is placed on page 23)

The fuzzy ontology efficiently employed for review and tweet classification. The main task is precise data collection, which can accelerate the fuzzy ontology development process. It contains the examined information such as "hospital" "is-a" feature of "medical centers"; "park ticket" "is-a" pass of "park"; "foods" "is-a" part of "restaurant." An ontology is a set of classes or concepts, properties (datatypes), instances, and relationships (object properties). A fuzzy concept of "park-ticket" is a concept where instances belong to a certain degree; hence "high-park-ticket" is a fuzzy concept because high is a blur predicate, and the concept is also blurry; thus, it should be considered a fuzzy concept. There are two types of relation; fuzzy object relations and fuzzy data type relations. Fuzzy object relations are used to link instances with a certain degree, such as "park-ticket has-rate high at a certain degree of 0.7." A fuzzy data type is used to assign an instance (e.g., "p-ticket has-fuzzy-value high") that includes the fuzzy predicate of ticket price. We gathered all information, such as medical centers, speed, cemeteries, parks, bridges, bus stations, tunnels, train stations, jails, sewage facilities, airports, streets, vehicles, crashes, accidents, and traffic, and delivered them to the fuzzy ontology. Fig. 3 shows our ontology classes, object properties, data properties, and fuzzy data types. These classes illustrate the concept of city features and transportation activity knowledge. Data and object properties define the relationships of the classes connected to the basic data types. The fuzzy data type shows the interval of membership variables. Sentiment analysis employs this fuzzy ontology-based semantic knowledge and evaluates pairs of city features and transportation activity polarity. In the last tasks, these feature polarity scores are gathered from all the tweets and reviews. The final sentiment analysis result and polarity values are obtained for transportation activities and city feature polarity.

(Fig. 3 is placed on page 18)

## 5. Sentiment analysis and polarity computation

In this section, the internal process of the proposed polarity computation system is introduced.

### 5.1. Tweet and review filtration

The proposed system uses various queries with ontology words to retrieve highly related tweets and reviews. Different queries are designed for tweet retrieval, and then, queries with more than 85 % recall are used. After the retrieval of tweets and reviews, the precision rate is very low. Therefore, a support vector machine (SVM) classifier is employed to identify related reviews and tweets, while unrelated ones are removed (Ali et al., 2016). N-gram techniques are employed to extract features from tweets. Bigram and trigram refer to extracting two and three adjacent feature words, respectively. These bigram and trigram features are presented in Fig 4. The proposed system employs specific functions to identify the value of each tweet. If a tweet value is greater than 0, that specifies that the tweet is related to a transportation or city feature; otherwise, the tweet is filtered out. A simple example is presented to explain the Fig. 4 classification tweets using the SVM classifier function: f (tweet or review) = 0.5 * Road + 0.6 * Accident + 0.1 * close − 0.3 * city name. The Fig. 4 tweet result is f (tweet or review) = 0.5 * 1 + 0.6 * 1 + 0.1 * 1 – 0.3 * 1= 0.9. If f (tweet or review) > 0, then it is a positive tweet or review (related to transportation and city); otherwise, the tweet or review is negative (Chen et al., 2013). After filtering, fuzzy ontology is applied to reviews and tweets to calculate the polarity of each feature. The detail is shown in Algorithm 1.

*5.2 Extraction of features and opinion words*

To more deeply explain the polarity computation, reviews for the city features parks and *New-York* were extracted to find their polarity, and then the overall city polarity. These reviews are, for example, "park is very clean, and the location is good", and "New York has a lot of facilities but is crowded." Before feature extraction, it is essential to eliminate stop words, articles (the, a, an) and prepositions (on, in, of) from reviews. After this, each sentence is checked to confirm that it is a complete clause with a noun and verb phrase. The conjunction *and* is used to separate the sentences in the above reviews. The first sentence in review 1 (Park is very clean and the location is good) and in review 2 (New-York has a lot of facility but crowded) has conjunctions and verbs. It should be split to classify complete clauses, which contain one noun, conjunction, and verb. The other sentences are composed of one noun and one verb and are recognized as complete sentences. The noun phrase of tweet or review is compared with fuzzy ontology classes (features). The matched feature with noun phrase shows that a given tweet or review is about a given feature. For example, *park* is a noun in the review sentence "Park is very clean" and can easily be identified. It shows that the review sentence is about park feature. The fuzzy ontology presents the concepts of transportation activities and city features and their relation to those concepts. The feature is extracted from the noun phrase in a single sentence. We imported feature extraction and polarity computation information into the fuzzy ontology. It is used to identify features in review sentences, such as accident, traffic, medical center, restaurant, hotel, and so on. Every extracted feature from tweets and reviews is compared with fuzzy ontology classes (features). The matched features are considered to predict polarity, and others are eliminated. Each feature has its own polarity words. For example, the feature park has clean and good opinion words with positive polarity values; *dusty* and *small* are opinion words with negative values, and *medium, normal,* and *average* are opinion words with neutral polarity values. Similarly, the opinion words of the feature New-York are *a lot* with a positive polarity value, and *crowded* with a negative polarity value.

**Algorithm 1:** Algorithm for polarity computation of given tweets and reviews
**Input:** Tweets and reviews of transportation activities and city features
**Output:** Polarity of transportation activities and city features
**Main Procedure:**
1. Retrieve tweets and reviews from social network
2. Eliminate needless content, such as special symbols (#, @, dates, etc.)
3. Split the tweets and reviews into sentences
4. Confirm that each sentence is a complete clause with noun and verb phrase
5. Remove stop words, and tag the sentences using POS tagging
6. Collect particular tweets and reviews using various queries
7. Identify features in noun phrases using ontology
8. Extract transportation activities and city features with sentiment words
9. Identify subjective and objective text of opinion words
10. Assign the SentiWordNet score to opinion words
11. Apply opinion lexicon for accurate sentiment analysis
12. Apply fuzzy ontology and SWRL rule for feature polarity
13. Compute the polarity of opinionated phrases
14. Return the sentiment and polarity value for a particular feature

*5.3 SentiWordNet scoring and polarity interval in the ontology*

SentiWordNet find the initial value of the corresponding opinion words. SentiWordNet is a lexical resource in which each word is connected with three numerical scores: positive, objective, and negative (Baccianella et al., 2010). These scores show how much the terms contained in the synset are positive, neutral, or negative. The opinion word *clean* is used as an adverb, and its SentiWordNet value is 0.5, which is its linguistic value in the ontology. The interval for each output is as follows: the strongly positive interval is [0.75-1], positive is [0.5-0.75], neutral is [0.5], negative is [0.25-0.5] and the strongly negative interval is [0.0-0.25]. This proposed system is based on these intervals, which identify the polarity by applying the input opinion word value. The proposed ontology uses various rules to compute the polarity and to find the causes of negative and strong negative polarity and traffic congestion problem as shown in Fig. 3. Some rules are explained in Table 1, where a reader can more properly understand the use of all the rules. In table 1 rules, OW stands for opinionated word. Similarly, SP, P, Neu, Neg and SN stand for strong positive, positive, neutral, negative and strong negative, respectively. Every opinion word uses SentiWordNet values, and then, assigns them to the fuzzy inference layer to asses the value of polarity terms: SP, P, Neu, Neg and SN. In this approach, various opinion words (verb, adverb, and adjective) are inputs, and the parameters are classic values connected with SentiWordNet such as the word *not* as an adverb and its parameter values are P: 0 O: 0.375 N: 0.625. Similarly, the words *noise* and *very* are verb and adjective, respectively, and their parameter values are P: 0 O: 1 N: 0 and P: 0.5 O: 0.375 N: 0.125, respectively.

*5.4 Polarity computation based on fuzzy inference layer*

The framework of the fuzzy inference layer is based on the knowledge and rule-base ontology as shown in Fig. 5. The fuzzy inference layer integrates the extracted opinionated phrases and their polarity values to verify the feature polarity for transportation and city. The fuzzy inference layer contains four components. Those are fuzzification, inference, knowledge base and rule base, and defuzzification. The fuzzification unit achieves the membership value of the opinion words. Let us consider the tweet sentences in Table 2 related to road and accident features. In these sentences, the opinion words for *Road* are {very, busy and closed}, and for *Accident*, they {horrible, closed, and killed}. SentiWordNet associates these sentiment words with special degrees, for example, very: 0.5, busy: 0.375, closed: 0.25, horrible: 0, killed: 0.125. The SWRL rules and linguistic values are stored in the ontology as explained in sections 3 and 4. Inference applies these rules to the fuzzy interval membership function (MF). The triangular MF is described to find the membership value for each input opinion word. There are five linguistic values (SP, P, Neu, Neg and SN) for each input variable, as shown in Fig. 5. The defuzzification converts the fuzzy output into normal terms, and provides the result in the form of a value, which is called the polarity value. This polarity value can be obtained through the following steps.

1. Apply the sentiment words to triangular MF for acquiring membership values as follows.

   Road {$\mu$(very) = 0.9, $\mu$(closed) = 1, $\mu$(busy) = 0.7 in N interval and 0.23 in Neu interval }

   Accident {$\mu$(horrible) = 1, $\mu$(closed) = 1, $\mu$(killed) = 3.5 in SN interval and 3.5 in N interval}

2. Find the input polarity (IP) value of road and accident opinion words using the rules as follows.

   Road:

   Rule1: If $\mu$(very) = 0.9, $\mu$(busy) = 0.23 or 0.7, $\mu$(closed) = 1 then IP is 0.25

   Accident:

   Rule2: If $\mu$(horrible) = 1, $\mu$(closed) = 1, $\mu$(killed) = 3.4 or 3.6 then IP value is 0

3. Select the minimum MF degree for the antecedents of road and accident opinion words and obtain the membership degree of consequent values using fitness function as follows.

   Road:

   Fitness [1] = min( $\mu$(very): 0.9, $\mu$(busy): 0.23, $\mu$(closed): 1) = 0.23

   Fitness [2] = min( $\mu$(very): 0.9, $\mu$(busy): 0.7, $\mu$(closed): 1) = 0.7

   Accident:

   Fitness [1] = min( $\mu$(horrible): 1, $\mu$(closed): 1, $\mu$(killed): 3.4) = 3.4

   Fitness [2] = min( $\mu$(horrible): 1, $\mu$(closed): 1, $\mu$(killed): 3.6) = 3.6

4. Find the final polarity of road and accident using equations 3 and 4.

$$\text{Output}[i] = \text{fitness}[i] * \text{IP}[i] \tag{3}$$

$$\text{Polarity} = \frac{\sum \text{output}[i] * \text{fitness}[i]}{\sum \text{fitness}[i]} \tag{4}$$

The road outputs 1 and 2 are equal to 0.057 and 0.175, respectively. Both accident outputs 1 and 2 are equal to 0. The polarity value for the feature road and accident sentences are 0.14 and 0, respectively. The intervals for each output are defined as follows: the strong positive interval is [0.75-1], positive is [0.5-0.75], neutral is [0.5], negative is [0.25-0.5] and strong negative is [0.0-0.25]. Based on these intervals, the polarity of both features road and accident is strong negative.

## 6. Experiment Results

To assess the usefulness of the proposed system, the proposed fuzzy ontology's performance was compared with classic ontology (Lau et al., 2009) and a simple fuzzy logic (RahmathP et al., 2014). Different types of search queries were composed to retrieve highly related, and the maximum number of tweets and reviews from e-commerce sites and Twitter. These tweets and reviews were stored in the database for further processing. The irrelevant reviews were filtered out, and every sentence confirmed to be in a valid format. This system retrieved 1404 and 1851 tweet and review sentences related to 7 different transportation features and 13 different city features, respectively. The average length of a tweet was 35 words, and the average length of a review was 56 words. The total number of transportation opinion words and city opinion words were 1101 and 1775, respectively. SentiWordNet-based raw datasets along with RDF (resource description framework) datasets was developed to compute the polarity. Both classical ontology and fuzzy ontology was built and compared their performance by using the same datasets. At first, a simple classic ontology was used to classify the tweets and reviews, predict polarities, and record the rate of precision, recall, and accuracy, and the function measure. Later, the fuzzy ontology was used to record the results. The precision and recall are basic measures for information retrieval efficiency. Precision (P) is the ratio of retrieved elements that are relevant and recall (R) is the ratio of relevant elements that are retrieved. The concepts are defined in table 3.

Mathematically, the rate of precision, recall, and accuracy, and the function measure can be calculated using the following equations (Cao et al., 2014).

$$\text{Precision (P)} = \frac{TP}{(TP + FP)} \times 100\% \tag{5}$$

$$\text{Recall (R)} = \frac{TP}{(TP + FN)} \times 100\% \tag{6}$$

$$\text{Accuracy (Ac)} = \frac{(TP + TN)}{(TP + FP + FN + TN)} \tag{7}$$

$$\text{Function Measure (FM)} = \frac{2^*P^*R}{P + R} \tag{8}$$

where "TP," "FP," "FN," and "TN" indicate "true positive," "false positive," "false negative," and "true negative" in the information retrieval, respectively. The comparative results of the experimental classic ontology and fuzzy ontology regarding transportation features' polarity are shown in Table 4.

This result shows that the proposed system performs well for the transportation features with explicit transportation aspects. The classic ontology-based system cannot learn these explicit features, because many implicit transportation features are used for different purposes. This illustrates that the proposed fuzzy ontology based system is more effective than the classic ontology. In addition, SWRL rules boost the performance efficiency in order to find the main cause for features of negative and strong negative polarity. The classic ontology system achieves a 73% average precision, whereas the proposed system succeeds 96% of the time, which is a 23% improvement. The main reason is that the fuzzy ontology can effectively classify sentiments related to transportation activity and city features. As a result, the accuracy of the sentiment analysis is high. In contrast, the classic ontology based system yields many noisy associated rules due to unrelated sentiment word retrieval that links unrelated polarity variables with features. Fig. 6 shows the precision, recall, accuracy and function measure results pertaining to different transportation features. It shows that the average precision, accuracy, and function measure significantly increased from 73% to 96%, 66% to 89%, and 71% to 94%, respectively, and recall incurred a 0.03% decrease. The results prove that more information that is precise can be extracted from feature tweet and review sentences using the fuzzy ontology.

The proposed fuzzy ontology based sentiment analysis was also employed to predict city-feature polarity. For the experimental classic ontology system, the semantic knowledge of city features is captured in a classic ontology to calculate the polarity of test comments. In particular, various equations (Polpinij et al., 2008) are used to find the polarity values of test comments, and then, equations 5, 6, 7 and 8 are applied to compute the overall results. To facilitate the comparison with the proposed system results, the SentiWordNet tool was used to assign the opinion value to sentiment words. If the opinion word is used for negative expressions, the value will be high as negative. Otherwise, the opinion word will get the high value as positive. Table 5 illustrates the precision, recall, accuracy, and function measure results achieved by using the classic ontology and the fuzzy ontology. The average improvements in the proposed system over the classic ontology system across 13 city features were 32%, 24%, and 25% in terms of precision, accuracy and function measure, respectively, whereas recall resulted in a 1.5% decrease. Fig. 7 clearly shows the performance of the proposed approach regarding a city feature opinionated phrase based on a crisp ontology and a fuzzy ontology.

(Table 5 is placed on page 27)

(Fig. 7 is placed on page 22)

It is observed that the proposed system performance increased significantly during sentiment analysis under the fuzzy ontology, while the classic ontology based system outperforms when the reviews or tweets are intensively blurred. The proposed system performance is more effective in both transportation and city feature sentiment analysis because of the fuzzy logic technique and SVM classifier. The fuzzy logic and feature-based sentiments are captured in the fuzzy ontology, and are then applied to drastically increase the precision of feature-level sentiment analysis. In addition, the average precision rate of the existing system was 74.1 %, whereas the proposed system achieves 97%. It proves that the average improvement of the proposed system over the existing system (Ali et al., 2016) is 23% in terms of precision. These analysis results confirm that the proposed fuzzy logic with an ontology-based system is more effective for transportation and city feature-level sentiment analysis.

## 7. Conclusion

This paper proposes a framework with fuzzy ontology based sentiment analysis of transportation and city feature reviews to facilitate ITSs and travelers. Various reasonable issues are considered, including feature extraction, opinion word extraction, the assertion of feature polarity value in the ontology, and polarity determination by utilizing fuzzy logic. The proposed technique offers a transportation system that determines a real-time traffic congestion mapping and provides travelers with a city feature opinion map of the city so they can plan before travelling. Indeed, the proposed technique successfully categorizes extremely obscure reviews, and intelligently determines transportation and city feature polarity. This system can automatically collect all the traffic data, and analyze these data to notify drivers before they reach risk areas, which save time and fuel. It can also improve transportation safety, solve traffic congestion problems in big cities, and provide travelers with comfortable traveling experiences in urban areas. Furthermore, this system can be connected to different information retrieval, text classification, and opinion-mining systems, since it can extricate a feature from unclear tweets and reviews, extract feature opinion words, and classify these feature opinion words into terms with more degrees of polarity. This work can be extended by enhancing the extraction procedure of true and false opinionated phrases in the field of sentiment analysis of transportation activity reviews.

Some future research works are debated as follows. Most of the proposed sentiment analysis ponder only polarity classification. However, in reality, reviews are intensively blurred. The proposed systems execute irrelevant reviews and consider the noun, verb, adjective, and adverb as sentiment words, which decrease the precision rate of sentiment analysis. Therefore, providing the required precision rate for effective sentiment analysis is complicated issue that must be addressed by using optimum mechanism.

Another future research direction that we propose is to develop sentiment analysis-based health prescription system. Indeed, such a system is needed to effectively measure the polarity degree of public health related tweets and to suggest precise drugs in limited time. In this system, the disease ontology-based domain knowledge with neural network can be considered to identify the diseases and to facilitate the information extraction of drugs from social network. For information extraction in opinion mining, we propose to use semi-supervised learning techniques that are more efficient to filter out irrelevant data.

**Acknowledgement**

This research was supported by a grant (12-TI-C01) from the Advanced Water Management Research Program funded by Ministry of Land, Infrastructure, and Transport of the Korean government.

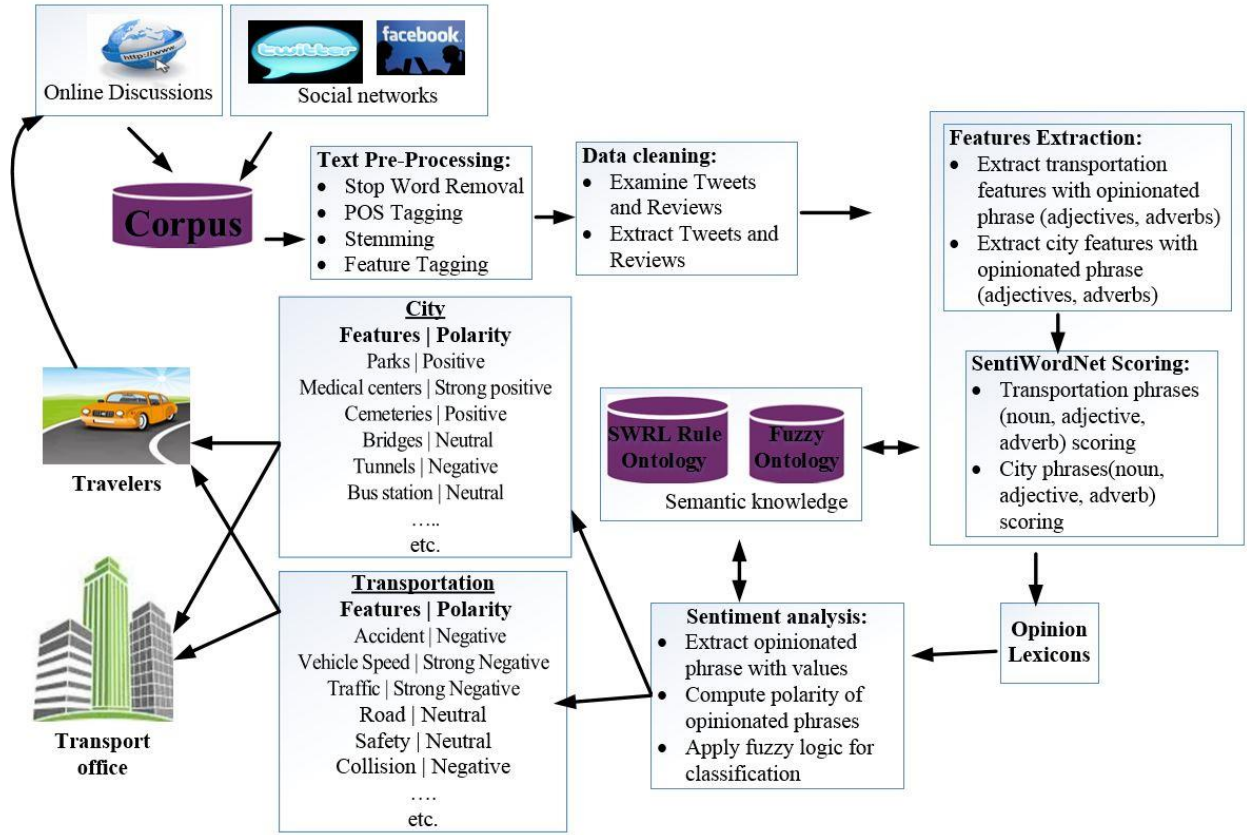

Fig. 1. Fuzzy semantic knowledge based transportation and city polarity system architecture.

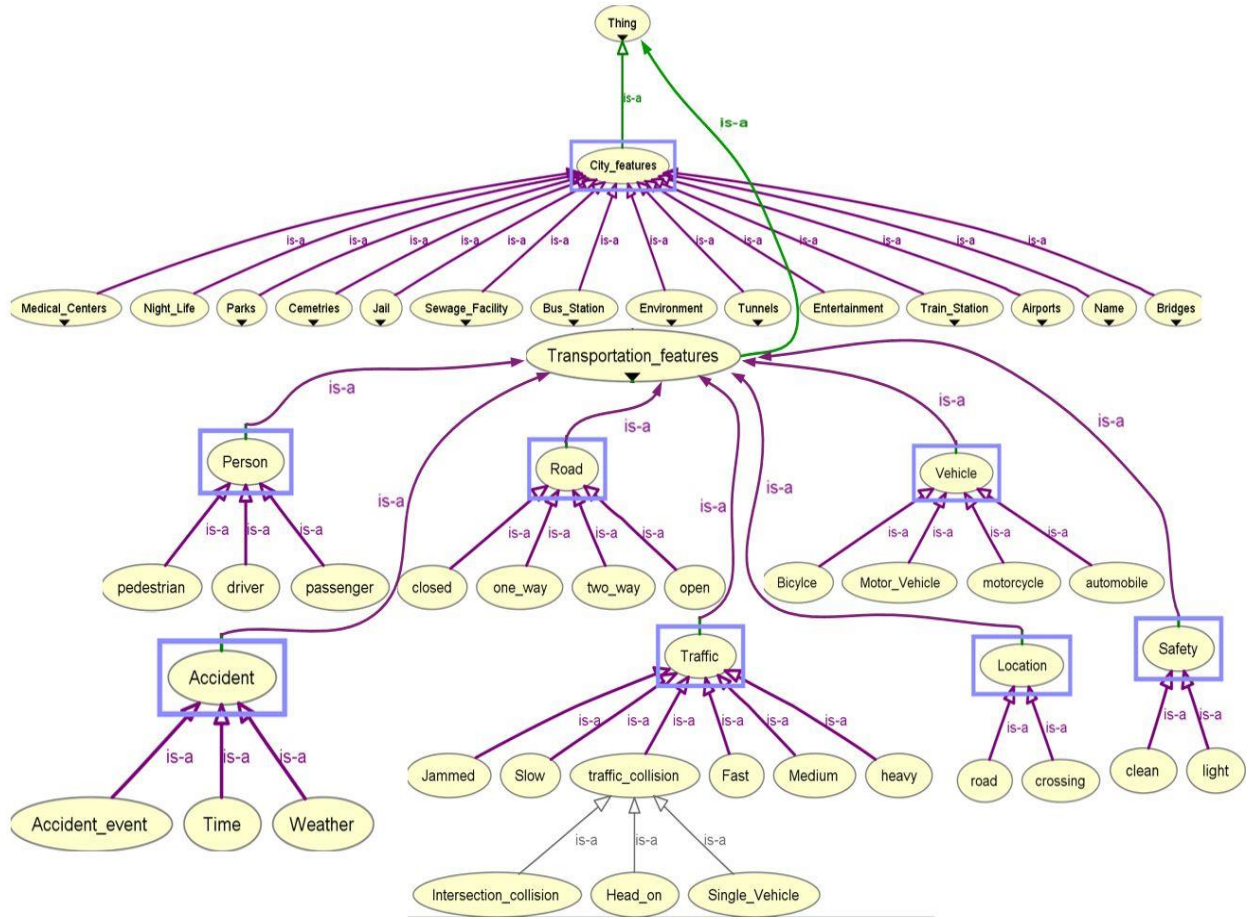

Fig. 2. Transportation and city features.

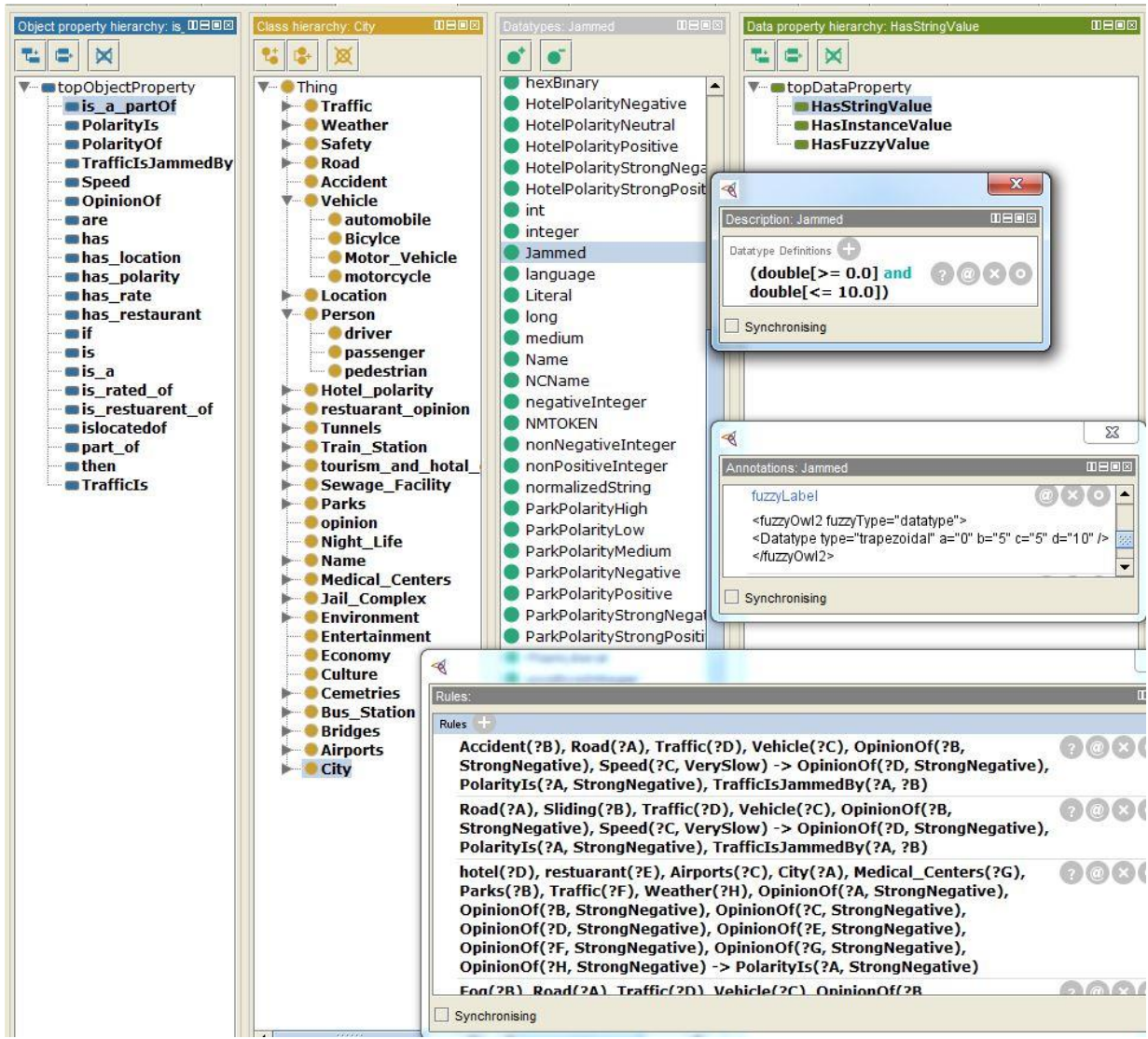

Fig. 3. Fuzzy ontology and rules.

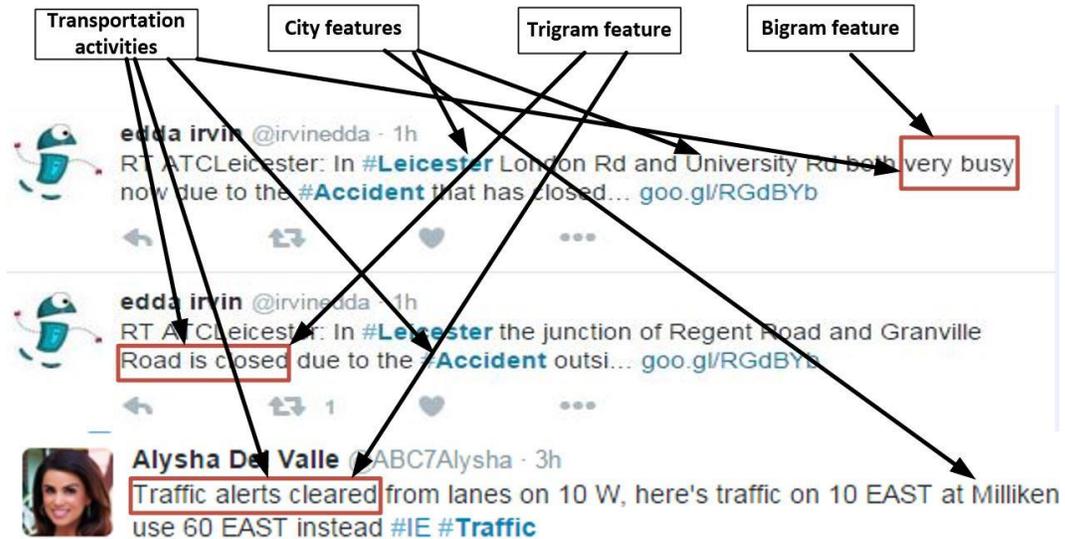

Fig. 4 Bigram and trigram tweet features.

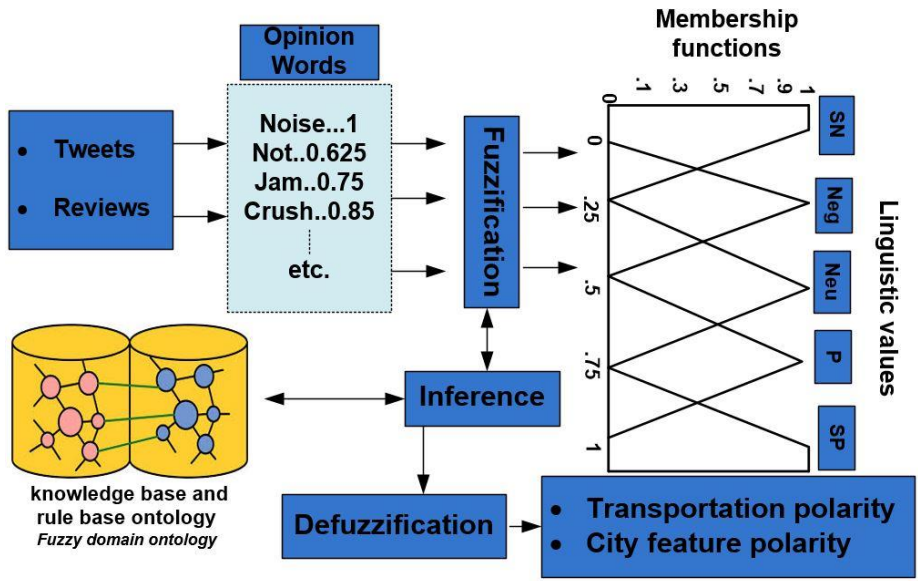

Fig. 5 Fuzzy inference layer for polarity computation.

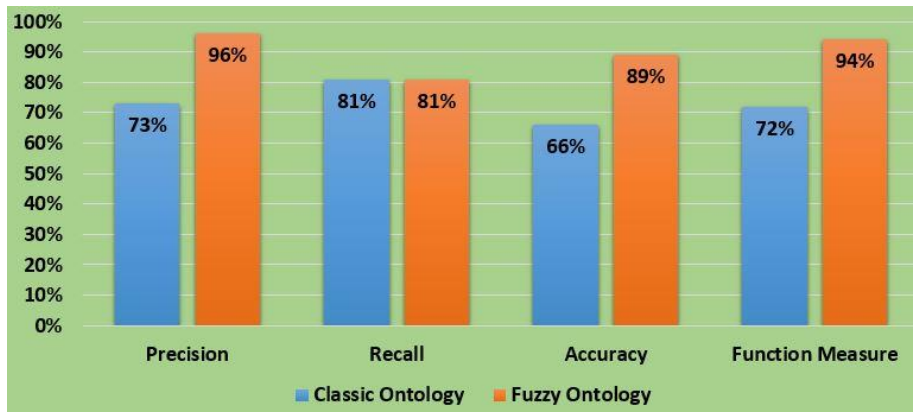

Fig. 6 Performance of the proposed approach regarding transportation features based on a classic ontology and a fuzzy ontology.

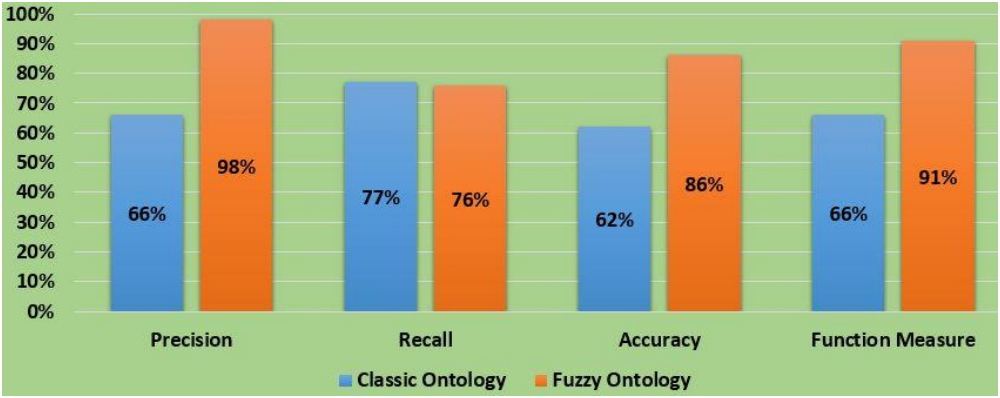
Fig. 7 Performance of the proposed approach regarding city features based on a classic ontology and a fuzzy ontology.

Table 1. Sample of ontology SWRL and IF-THEN rules for polarity computation.

| Rules | Explanation |
|---|---|
| Accident(?B), Road(?A), Traffic(?D), Vehicle(?C), OpinionOf(?B, SN), Speed(?C, VerySlow) -> OpinionOf(?D, SN), PolarityIs(?A, SN), TrafficIsJammedBy(?A, ?B) | If the speed of 'vehicle' is 'very low' and opinion of the 'accident' is 'strong negative' then the 'traffic' is jammed because of 'accident'. Thus, the 'traffic' and 'road' polarity will be counted as 'strong negative'. |
| Accident(?B), Road(?A), Traffic(?D), Vehicle(?C), OpinionOf(?B, Neg), Speed(?C, VerySlow) -> OpinionOf(?D, N), PolarityIs(?A, Neg), TrafficIsJammedBy(?A, ?B) | If the opinion of 'accident' is 'negative' and the speed of 'vehicle' is 'very slow' then the 'traffic' is jammed because of 'accident'. Therefore, the 'traffic' and 'road' polarity will be counted as 'negative'. |
| Accident(?B), Road(?A), Traffic(?D), Vehicle(?C), OpinionOf(?B, Neu), Speed(?C, VerySlow) -> OpinionOf(?D, N), PolarityIs(?A, Neu ), TrafficIsJammedBy(?A, ?C) | If the opinion of 'accident' is 'neutral' and the speed of 'vehicle' is 'very slow' then the 'traffic' is jammed because of 'vehicles'. Thus, the 'traffic' and 'road' polarity will be counted as 'negative' and 'neutral', respectively. |
| Accident(?B), Road(?A), Traffic(?D), Vehicle(?C), OpinionOf(?B, P), Speed(?C, VerySlow) -> OpinionOf(?D, P), PolarityIs(?A, Neu ), TrafficIsJammedBy(?A, ?C) | If the opinion of 'accident' is 'neutral' and the speed of 'vehicle' is 'very slow' then the 'traffic' is jammed because of 'vehicles'. As a result, the 'traffic' and 'road' polarity will be counted as 'negative' and 'neutral', respectively. |
| Accident(?B), Road(?A), Traffic(?D), Vehicle(?C), OpinionOf(?B, SP), Speed(?C, Slow) -> OpinionOf(?D, Neu), PolarityIs(?A, Neu ), TrafficIsJammedBy(?A, ?C) | If the opinion of 'accident' is 'strong positive' and the speed of 'vehicle' is 'slow' then the 'traffic' is jammed because of 'vehicles'. Hence, the 'traffic' and 'road' polarity will be counted as 'neutral'. |
| Accident(?B), Road(?A), Traffic(?D), Vehicle(?C), OpinionOf(?B, SP), Speed(?C, Normal) -> OpinionOf(?D, P), PolarityIs(?A, Neu ) | If the opinion of 'accident' is 'strong positive' and the speed of 'vehicle' is 'normal' then the 'traffic' opinion and the 'road' polarity will be counted as 'positive' and 'neutral', respectively. |
| Accident(?B), Road(?A), Traffic(?D), Vehicle(?C), OpinionOf(?B, SP), Speed(?C, fast) -> OpinionOf(?D, P), PolarityIs(?A, P) | If the opinion of 'accident' is 'strong positive' and the speed of 'vehicle' is 'fast' then both the 'traffic' opinion and the 'road' polarity will be counted as 'positive'. |
| Hotel(?D), Restuarant(?E), City(?A), Medical_Centers(?G), Airports(?C), Parks(?B), Weather(?H), Traffic(?F), OpinionOf(?A, SN), OpinionOf(?B, SN), OpinionOf(?C, SN), OpinionOf(?D, SN), OpinionOf(?E, SN), OpinionOf(?F, SN), OpinionOf(?G, SN), OpinionOf(?H, SN) -> PolarityIs(?A, SN) | Illustrates that the city polarity will be 'strong negative' if the polarity of all city features is 'strong negative'. |
| If Road_OW (Adj/verb/adverb/obj) is SN, Vechicle_OW (Adj/verb/adverb/obj) is SN, Loc_OW (Adj/verb/adverb/obj) is SN, Accident_OW (Adj/verb/adverb/obj) is SN, Traffic_OW (Adj/verb/adverb/obj) is SN, Safety_OW (Adj/verb/adverb/obj) is SN then polarity is SN | If the opinionated words (Adj/verb/adverb/obj) of 'road', 'vehicle', 'location', 'accident', 'traffic', and 'safety' are 'strong negative' then polarity is 'strong negative'. |
| If Road_OW (Adj/verb/adverb/obj) is Neg, Vechicle_OW (Adj/verb/adverb/obj) is N, Loc_OW (Adj/verb/adverb/obj) is N, Accident_OW (Adj/verb/adverb/obj) is Neg, Traffic_OW (Adj/verb/adverb/obj) is Neg, Safety_OW (Adj/verb/adverb/obj) is Neg then polarity is Neg | If the opinionated words (Adj/verb/adverb/obj) of 'road', 'vehicle', 'location', 'accident', 'traffic', and 'safety' are 'negative' then polarity is 'negative'. |
| If Road_OW (Adj/verb/adverb/obj) is Neu, Vechicle_OW (Adj/verb/adverb/obj) is Neu, Loc_OW (Adj/verb/adverb/obj) is Neu, Accident_OW (Adj/verb/adverb/obj) is Neu, Traffic_OW (Adj/verb/adverb/obj) is Neu, Safety_OW (Adj/verb/adverb/obj) is Neu then polarity is Neu | If the opinionated words (Adj/verb/adverb/obj) of 'road', 'vehicle', 'location', 'accident', 'traffic', and 'safety' are 'neutral' then polarity is 'neutral'. |
| If Road_OW (Adj/verb/adverb/obj) is P, Vechicle_OW (Adj/verb/adverb/obj) is P, Loc_OW (Adj/verb/adverb/obj) is P, Accident_OW (Adj/verb/adverb/obj) is P, Traffic_OW (Adj/verb/adverb/obj) is P, Safety_OW (Adj/verb/adverb/obj) is P then polarity is P | If the opinionated words (Adj/verb/adverb/obj) of 'road', 'vehicle', 'location', 'accident', 'traffic' and 'safety' are 'positive' then polarity is 'positive'. |
| If Road_OW (Adj/verb/adverb/obj) is SP, Vechicle_OW (Adj/verb/adverb/obj) is SP, Loc_OW (Adj/verb/adverb/obj) is SP, Accident_OW (Adj/verb/adverb/obj) is SP, Traffic_OW (Adj/verb/adverb/obj) is SP, Safety_OW (Adj/verb/adverb/obj) is SP then polarity is SP | If the opinionated words (Adj/verb/adverb/obj) of 'road', 'vehicle', 'location', 'accident', 'traffic', and 'safety' are ' strong positive' then polarity is 'strong positive'. |

Table 2. Tweet sentences.

| |
|---|
| Road Quezon is very busy! Want help to get to my home. |
| Road is closed due to accident. |
| Just saw a horrible accident on road in Quezon. Please be careful, life is important. |
| 3 killed in Quezon road accident. |

Table 3. Performance evaluation of the information retrieval.

| Information Retrieval | Relevant | Irrelevant |
|---|---|---|
| Retrieved | TP | FP |
| Not Retrieved | FN | TN |

Table 4. Transportation feature polarity results based on a classic ontology and the proposed system.

| Transportation activities or features | Total retrieved tweet and review sentences | Classic Ontology | | | | Fuzzy Ontology | | | |
|---|---|---|---|---|---|---|---|---|---|
| | | P | R | Ac | FM | P | R | Ac | FM |
| Road | 134 | 61.52 | 73.22 | 58.72 | 62.36 | 98.08 | 77.52 | 87.60 | 92.50 |
| Accident | 231 | 65.03 | 83.81 | 60.19 | 69.14 | 96.76 | 84.47 | 91.98 | 95.60 |
| Vehicle | 155 | 78.99 | 76.43 | 67.05 | 72.17 | 98.73 | 80.47 | 89.89 | 94.42 |
| Traffic | 212 | 74.49 | 85.52 | 67.98 | 75.00 | 96.14 | 87.43 | 94.19 | 96.78 |
| Safety | 190 | 77.01 | 76.56 | 67.14 | 71.40 | 93.38 | 80.07 | 85.63 | 91.70 |
| Location | 182 | 79.67 | 92.70 | 75.60 | 81.16 | 93.77 | 78.27 | 85.44 | 90.94 |
| Person | 100 | 73.02 | 81.13 | 64.33 | 72.06 | 96.15 | 80.91 | 88.33 | 93.46 |
| Average | | 72.82 | 81.34 | 65.86 | 71.90 | 96.14 | 81.30 | 89.01 | 93.63 |

Table 5. City feature polarity results based on a classic ontology and the proposed system.

| City features | Total retrieved tweet and review sentences | Classic Ontology | | | | Fuzzy Ontology | | | |
|---|---|---|---|---|---|---|---|---|---|
| | | P | R | Ac | FM | P | R | Ac | FM |
| Medical_ Centers | 134 | 75.64 | 73.90 | 70.60 | 69.28 | 98.62 | 81.06 | 91.04 | 94.69 |
| Nightlife | 131 | 75.00 | 67.58 | 56.09 | 65.15 | 98.39 | 84.01 | 93.30 | 96.15 |
| Parks | 255 | 65.75 | 76.51 | 64.88 | 66.13 | 97.20 | 78.41 | 87.02 | 92.60 |
| Cemeteries | 112 | 54.76 | 78.66 | 44.34 | 60.93 | 98.32 | 59.64 | 75.00 | 81.53 |
| Jail | 90 | 62.73 | 96.52 | 62.08 | 68.94 | 97.83 | 70.84 | 80.56 | 88.52 |
| Sewage facility | 182 | 79.67 | 79.50 | 71.25 | 74.24 | 97.51 | 79.80 | 88.11 | 93.49 |
| Bus station | 100 | 73.61 | 87.95 | 74.60 | 75.72 | 99.73 | 69.53 | 80.80 | 88.49 |
| Environments | 161 | 49.42 | 68.36 | 55.07 | 53.52 | 99.08 | 78.52 | 88.28 | 93.51 |
| Tunnels | 151 | 56.25 | 82.97 | 62.95 | 63.53 | 94.38 | 65.12 | 76.49 | 83.66 |
| Entertainment | 178 | 53.89 | 78.87 | 64.02 | 60.47 | 97.12 | 74.38 | 85.05 | 90.30 |
| Train Station | 95 | 63.64 | 74.54 | 52.11 | 64.08 | 100.00 | 84.79 | 94.87 | 97.33 |
| Airports | 167 | 74.19 | 78.22 | 64.31 | 71.08 | 99.04 | 79.63 | 89.37 | 94.10 |
| Bridges | 95 | 77.27 | 71.82 | 67.37 | 68.69 | 100.00 | 80.00 | 90.53 | 94.74 |
| Average | | 66.29 | 77.34 | 62.28 | 66.29 | 98.25 | 75.82 | 86.19 | 91.47 |